\documentclass[aps,reprint]{revtex4-1}

\usepackage{main}

\begin{document}
\title{Designing spontaneous behavioral switching via chaotic itinerancy}

\author{Katsuma Inoue}
\email{k-inoue@isi.imi.i.u-tokyo.ac.jp}
\affiliation{Graduate School of Information Science and Technology, The University of Tokyo, 7-3-1 Hongo, Bunkyo-ku, Tokyo 113-8656, Japan}

\author{Kohei Nakajima}
\email{k_nakaijma@mech.t.u-tokyo.ac.jp}
\affiliation{Graduate School of Information Science and Technology, The University of Tokyo, 7-3-1 Hongo, Bunkyo-ku, Tokyo 113-8656, Japan}

\author{Yasuo Kuniyoshi}
\email{kuniyosh@isi.imi.i.u-tokyo.ac.jp}
\affiliation{Graduate School of Information Science and Technology, The University of Tokyo, 7-3-1 Hongo, Bunkyo-ku, Tokyo 113-8656, Japan}
\date{\today}

\begin{abstract}
Chaotic itinerancy is a frequently observed phenomenon in high-dimensional and nonlinear dynamical systems, and it is characterized by the random transitions among multiple quasi-attractors.
Several studies have revealed that chaotic itinerancy has been observed in brain activity, and it is considered to play a critical role in the spontaneous, stable behavior generation of animals.
Thus, chaotic itinerancy is a topic of great interest, particularly for neurorobotics researchers who wish to understand and implement autonomous behavioral controls for agents.
However, it is generally difficult to gain control over high-dimensional nonlinear dynamical systems.
Hence, the implementation of chaotic itinerancy has mainly been accomplished heuristically.
In this study, we propose a novel way of implementing chaotic itinerancy reproducibly and at will in a generic high-dimensional chaotic system.
In particular, we demonstrate that our method enables us to easily design both the trajectories of quasi-attractors and the transition rules among them simply by adjusting the limited number of system parameters and by utilizing the intrinsic high-dimensional chaos.
Finally, we quantitatively discuss the validity and scope of application through the results of several numerical experiments.
\end{abstract}

\maketitle

\section{Introduction}
Designing a cognitive architecture that acts spontaneously in the real-world environment is one of the ultimate goals in the field of cognitive robotics \cite{pfeifer2001understanding}.
A cognitive agent is expected to have autonomy; i.e., the agent should behave independently of the designer's control while maintaining its identity.
Furthermore, adaptability is another requirement for cognitive functionality.
Thus, the agent must select the appropriate behavior continuously and robustly in response to the changing environment in real-time.
To summarize, the agent's cognitive behavior should be implemented through the body-environment interaction while still enabling the agent to maintain its autonomy and adaptability.

In the conventional context of robotics and artificial intelligence, designers often take top-down approaches to provide an agent with a hierarchical structure corresponding to the behavioral category.
This representation-based approach has a critical limitation in the design of a cognitive agent: the static and one-to-one relationship between the behavior and structure makes it difficult to adapt flexibly to the dynamically changing environment and developing body.
For example, it has been considered that the motion control systems of living things, including humans, realize their high-order motion plans by combining the reproducible motor patterns called \textit{motion primitives} \cite{flash2005motor}.
Inspired by this viewpoint, we tend to realize an agent's behavior control with a predetermined static hierarchical structure.
However, such a hierarchical structure does not exist in living organisms from the beginning; rather, these structures are cultivated through the body's development and dynamic interactions with the environment.
Thus, it is important to introduce a dynamical perspective to understand a hierarchical structure of behavior control generated in animals that possesses adaptability and flexible plasticity.

In robotics, approaches based on dynamical systems theory have been applied to analyze and control agents being modeled as sets of variables and parameters on a phase space \cite{beer1995dynamical,jaeger1995identification}.
This \textit{dynamical systems approach} can deal with both the functional hierarchy and the elementary motion in a unified form by expressing the whole physical constraints of the agent as the temporal development of state variables, namely \textit{dynamics}.
For example, Jaeger \cite{jaeger1995identification} sketched a pioneering idea of an algorithm where the behavior of an agent is expressed as dynamics, and both the behavioral regularity (referred to as \textit{transient attractors}) and the higher-order relationships among them are extracted in a bottom-up manner.
Therefore, the dynamical systems approach has the potential to model an agent's hierarchical behavior as the dynamics of interaction without a top-down structure given by the external designer.

Following this dynamical systems perspective, \textit{chaotic itinerancy} (CI) \cite{ikeda1989maxwell,kaneko1990clustering,tsuda1991chaotic} is a powerful option for modeling spontaneous behavior with functional hierarchy.
CI is a frequently observed, nonlinear phenomenon in high-dimensional dynamical systems, and it is characterized by random transitions among locally contracting domains, namely \textit{quasi-attractors}.
In general, a chaotic system has an initial sensitivity, and a slight difference in the phase space is exponentially expanded in a certain direction with temporal development.
Conversely, multiple transiently predictable dynamics can be repeatedly observed in a chaotic system, yielding CI despite the global chaoticity and initial sensitivity.
Interestingly, this type of hierarchical dynamics frequently emerges from high-dimensional chaos even without hierarchical mechanisms, implying that explicit structure is not necessarily needed for implementing hierarchical behaviors.
Furthermore, the chaoticity plays an important role in forming the autonomy of an agent as it is virtually impossible for the designer to predict and control an agent's behavior completely due to the agent's initial sensitivities, which essentially ensures the agent's independence from the designer.
Thus, CI would work as an effective tool for implementing the intellectual behavior of a cognitive agent by embedding the behavior in the form of a quasi-attractor and maintaining the autonomy of the agent with the chaoticity.

Historically, CI was first found in a model of optical turbulence \cite{ikeda1989maxwell}.
Since its discovery, similar phenomena have been numerically obtained in various setups \cite{tsuda1987memory,kaneko1990clustering,adachi1997associative}.
Several physiological studies have reported that CI-like dynamics have even occurred in brain activity, suggesting that CI plays an essential role in forming cognitive functions \cite{tsuda2001toward,tsuda2015chaotic}.
For example, Freeman et al. \cite{freeman1987simulation} revealed that an irregular transition among learned states was observed in the electroencephalogram pattern of a rabbit olfactory when a novel input was given, indicating that the cognitive conditions corresponding to ``I don't know'' are internally realized as CI-like dynamics.
Furthermore, a recent observation of rat auditory cortex cell activity revealed the existence of a random shift among different stereotypical activities corresponding to individual external stimuli during anesthesia \cite{luczak2009spontaneous}.
Based on these reports, Kurikawa et al. \cite{kurikawa2013embedding} suggested the novel idea of ``memory-as-bifurcation'' to understand the mechanism of the transitory phenomenon.
They reproduced it in an associative memory model in which several input-output functions were embedded by Hebbian learning.
An intermittent switching among meta-stable patterns was also observed in a recurrent neural network (RNN) by installing multiple feedback loops trained to output a specific transient dynamic corresponding to external transient inputs \cite{suetani2019multiple}.
CI-like dynamics arise not only in nervous systems but also in interactions between agent's bodies and their surrounding environments \cite{kuniyoshi2004dynamic,kuniyoshi2006early,ikegami2007simulating,park2017chaotic}.
Kuniyoshi et al. \cite{kuniyoshi2006early} developed a human fetal development model, for example, by coupling chaotic central pattern generators and a musculoskeletal system, and they reported that several common behaviors, such as crawling and rolling over, emerge from the physical constraint.
Therefore, CI is a nonlinear phenomenon of high-dimensional dynamical systems and is thought to play a significant role in generating structural behavior.

Inspired by the contribution of CI to the cognitive functions and spontaneous motion generation of agents, CI has been utilized for motion control in the field of neurorobotics and cognitive robotics by designing the CI trajectory.
For example, Namikawa and Tani \cite{namikawa2008model,namikawa2010learning,namikawa2011neurodynamic} designed stochastic motion-switching among predetermined motor primitives in a humanoid robot by using a hierarchical, deterministic RNN controller.
In this study, it was confirmed that lower-order RNNs with larger time constants stably produced the trajectories of motion primitives, whereas higher-order RNNs with larger time constants realized a pseudo-stochastic transition by exploiting self-organized chaoticity.
Steingrube et al. \cite{steingrube2010self} designed a robot that skillfully broke a deadlock state in which the motion had completely stopped by employing chaos in the RNN controller.
Hence, it can be interpreted that CI-like dynamics were embedded in the coupling of the body and the surrounding environment.

While CI is a fascinating phenomenon in high-dimensional dynamical systems, roboticists also find it a useful tool for designing an agent's behavior structure while maintaining the agent's autonomy.
However, it has generally been difficult to embed desired quasi-attractors at will due to their nonlinearity and high dimensionality.
For example, in Namikawa and Tani's method \cite{namikawa2008model,namikawa2010learning,namikawa2011neurodynamic}, the internal connections of an RNN was trained with backpropagation through time (BPTT) \cite{werbos1990backpropagation}; however, embedding a long-term input-output function in an RNN by the gradient descent method is generally unstable and requires a large number of the learning epochs, as pointed out in \cite{bengio1994learning}.
Furthermore, their method required both a hierarchical structure and the same number of separated modules as the motor primitives, restricting the scalability and the range of its application.
Methods using the associated memory model \cite{adachi1997associative,oku2011associative} are also unsuitable for designing output with complicated spatiotemporal patterns because the embedded quasi-attractors are limited to fixed-point attractors.

\begin{figure*}[hbtp]
  \centering
  \includegraphics[width=0.8\textwidth]{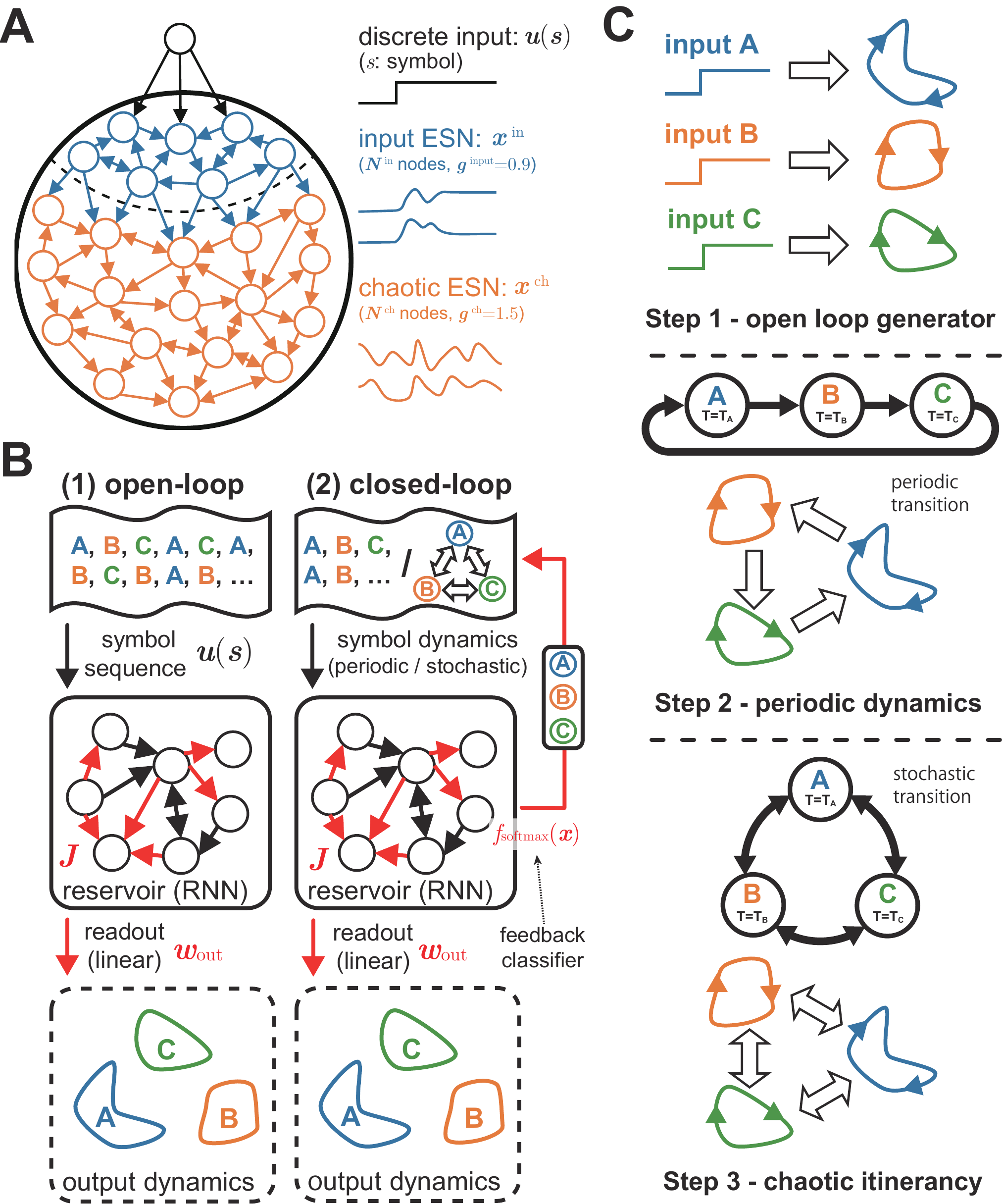}
  \caption[fig1]{
    Experimental setups.
    (A) Schematic diagram of a high-dimensional chaotic system prepared in our experiments.
    The system can be divided into two parts: input ESN and chaotic ESN.
    Input ESN acts as an interface between the discrete input and the chaotic ESN, generating transient dynamics projecting onto the chaotic ESN when the symbol input switches.
    To prevent the chaotic ESN from becoming non-chaotic due to the bifurcation, the connection between the input ESN and the chaotic ESN is trained to output transient dynamics converging to $0$ (see the Appendix for the detailed information about the transient dynamics).
    (B) Two experimental schemes.
    In the open-loop scheme, the symbol input is externally given.
    On the other hand, in the closed-loop one, the symbol input is autonomously generated by the additional feedback loop.
    In our method, we change the elements represented by red arrows to embed desired CI dynamics.
    (C) Outline diagrams of our batch learning methods composed of a three-step procedure.
    In step 1, the parameters of the network and readout are trained to output the quasi-attractors and the output dynamics corresponding to the symbols.
    In step 2 and step 3, the symbol sequence is autonomously yielded.
    We prepare periodic symbol transition patterns as the target in step 2 and stochastic symbol transition rules in step 3.
  }\label{fig:1}
\end{figure*}

In this study, we propose a novel algorithm, freely designing both the trajectories of quasi-attractors and transition rules among them in a generic setup of high-dimensional chaotic dynamical systems.
Our method employs batch learning composed of the following three-step procedure (\cref{fig:1}):

\begin{enumerate}[label=\textit{Step \arabic*}:]
\item
Prepare a high-dimensional chaotic system and modify the interactions (internal parameters) so that the system reproducibly generates intrinsic chaotic trajectories (\textit{innate trajectories}) corresponding to the type of the discrete inputs (named \textit{symbol}).
At the same time, alter the linear regression model (named \textit{readout}) to output the designated trajectories (\textit{output dynamics}) by exploiting the high dimensionality and nonlinearity of the internal dynamics.
Our method can be applied to a wide range of chaotic dynamical systems since neither modules nor hierarchical structures are required.
Also, this embedding process is accomplished by modifying fewer parameters, utilizing the method of reservoir computing (RC) \cite{maass2002real,jaeger2004harnessing}.
Therefore, our scheme is more stable and less computationally expensive than the conventional methods using backpropagation to train the network parameters.

\item
Add a feedback classifier to the trained chaotic systems for generating specific symbol dynamics autonomously.
In the training of the feedback discriminator, the network's internal parameters are fixed, as with the readout in step 1.
Thus, by making the most of the information processing of the innate trajectory, the feedback discriminator achieves multiple symbol transition rules with minimum additional computational capacity (i.e., nonlinearity and memory).

\item
Regulate the feedback unit added in step 2 to design designated stochastic symbol transition rules.
The deterministic system is expected to imitate the stochastic process by employing intrinsic chaoticity.
The system repeatedly generates the quasi-attractors embedded in step 1 in synchronization with the pseudo-stochastic symbol transition, meaning that the design of the desired CI dynamics is completed.

\end{enumerate}
In this study, we demonstrate that the trajectories of quasi-attractors and their transition rules can be designed using the three steps described above.
In step 1, we show that the desired output dynamics can be designed with high operability by utilizing the embedded internal dynamics reproducibly generated after the innate training.
Next, in step 2, we demonstrate that various types of periodic symbol sequences switching at a certain interval can be implemented simply by adjusting the parameters of a feedback loop attached to the system.
Finally, in step 3, we prepare several stochastic symbol transition rules governed by a finite state machine and show that the system can simulate these stochastic dynamics by making use of the system's chaoticity.
We also discuss the proposed method's validity and adaptability through several numerical experiments.

\section{Proposed method}
\subsection{System architecture}
In our method, we aimed to embed $M$ types of quasi-attractors and the transition rules among them in an RNN.
We prepared $M$ discrete symbols $s\in S~(S:=\{s_1, s_2, \cdots s_M\})$.
Each symbol corresponds to each individual quasi-attractor.
We used echo state network (ESN) \cite{jaeger2001echo}, one type of RNN, as a high-dimensional chaotic system.
As shown in \cref{fig:1}A, we prepared a generic RNN composed of a non-chaotic \text{input ESN} ($N^{\text{in}}$ nodes) working as an input transient generator as well as a \textit{chaotic ESN} ($N^{\text{ch}}$ nodes) yielding chaotic dynamics.
The dynamics of input ESN $\bm{x}^{\text{in}}(t) \in \Real^{N^{\text{in}}}$ and chaotic ESN $\bm{x}^{\text{ch}}(t) \in \Real^{N^{\text{in}}}$ are given as the following differential equations:
\begin{align}
    \tau \diff{\bm{x}^{\text{in}}}{t}{}(t) &= -\bm{x}^{\text{in}}(t) + \tanh \left(g^{\text{in}} J^{\text{in}} \bm{x}^{\text{in}}(t) + \bm{u}^{\text{in}}(s(t)) \right) \label{equ:input_ESN} \\
    \tau \diff{\bm{x}^{\text{ch}}}{t}{}(t) &= -\bm{x}^{\text{ch}}(t) + \tanh \left(g^{\text{ch}} J^{\text{ch}} \bm{x}^{\text{ch}}{}(t) + J^{\text{ic}}\bm{x}^{\text{in}}(t) \right) \label{equ:chaotic_ESN}
\end{align}
where $\tau \in \Real$ is a time constant,
$\tanh$ element-wise hyperbolic tangent,
$g^{\text{in}}, g^{\text{ch}} \in \Real$ are scaling parameters,
$u^{\text{in}}(s) \in \Real^{N^\text{in}}$ is discrete input projected onto input ESN when symbol $s$ is given,
$J^{\text{in}} \in \Real^{N^{\text{in}}\times N^{\text{in}}}, J^{\text{ch}}\in \Real^{N^{\text{ch}}\times N^{\text{ch}}}$ are connection matrices,
and $J^{\text{ic}}\in \Real^{N^{\text{ch}}\times N^{\text{in}}}$ is a feed-forward connection matrix between input ESN and chaotic ESN.
Each element of $J^{\text{in}}$ is sampled from a normal distribution $\mathcal{N}\left(0,\frac{1}{N^{\text{in}}}\right)$.
$J^{\text{ch}}$ is a random sparse matrix with density $p=0.1$ whose elements are also sampled from a normal distribution $\mathcal{N}\left(0,\frac{1}{pN^{\text{ch}}}\right)$.
We used $\tau=10.0, g^{\text{in}}=0.9, g^{\text{ch}}=1.5$ to make input ESN non-chaotic and chaotic ESN chaotic \cite{sompolinsky1988chaos}。
Also, to prevent chaotic ESN from becoming non-chaotic due to the bifurcation caused by the strong bias term, we tuned $J^{\text{ic}}$ before hand to project transient dynamics converging to $0$ onto the chaotic ESN when the same symbol input continues to be given (see the Appendix for detailed information about the transient dynamics).
In any case, the whole RNN dynamics $\bm{x}(t)\in\Real^{N^\text{in}+N^\text{ch}}$ concatenating \cref{equ:input_ESN,equ:chaotic_ESN} can be represented by the following single equation ($\odot$ represents an element-wise product):
\begin{align}
    \tau \diff{\bm{x}}{t}{}(t) &= -\bm{x}(t) + \tanh \left(\bm{g}\odot(J\bm{x}(t)) + \bm{u}(s(t)) \right)
\end{align}
where $\bm{x},\bm{g},J,\bm{u}$ are defined by the following equations:
\begin{align}
    \bm{x}(t) &:= [\bm{x}^{\text{in}}(t); \bm{x}^{\text{ch}}(t)] \\
    \bm{g} &:= [
        \underbrace{g^{\text{in}}, \cdots g^{\text{in}}}_{N^{\text{in}}}
        \underbrace{g^{\text{ch}}, \cdots g^{\text{ch}}}_{N^{\text{ch}}}
    ]^T \\
    J &:= \begin{bmatrix}
        J^{\text{in}} & \bm{0} \\
        J^{\text{ic}} & J^{\text{ch}}
    \end{bmatrix} \\
    \bm{u}(s) &:= [\bm{u}^{\text{in}}(s); \bm{0}]
\end{align}
The output dynamics are calculated by the linear transformation of the internal dynamics $\bm{x}(t)$, that is, the linear readout $w_\text{out}\in\Real^{N^{\text{in}}+N^{\text{ch}}}$ is trained to approximate the following target dynamics $f_{\text{out}}(t)$:
\begin{align}
    \bm{w}_{\text{out}}^T\bm{x}(t) \approx f_{\text{out}}(t)
\end{align}
The symbol dynamics $s(t)$ itself, which is externally given in step 1, is finally generated autonomously with a closed-loop system (\cref{fig:2}B(2)).
In the feedback loop, the following softmax classifier $f_{\text{softmax}}:\Real^{N^{\text{in}}+N^{\text{ch}}}\to S$ is attached:
\begin{align}
    f_{\text{softmax}}\left(\bm{x}(t)\right):= \argmax_{s\in S} \bm{w}_{s}^T\bm{x}(t)
\end{align}
where $\bm{w}_{s}\in \Real^{N^{\text{in}}+N^{\text{ch}}}$ represents the connection matrix whose elements are trained to approximate designated symbol dynamics $s(t)$ (i.e., $s(t)\approx f_{\text{softmax}}(\bm{x}(t))$).

To summarize, we designed the desired quasi-attractors, output dynamics, and symbol dynamics by tuning the parameters of the RNN connections $J$, the readout $\bm{w}_{\text{out}}$, and the softmax classifier $\bm{w}_s$, respectively.

\subsection{First-order-reduced and controlled-error (FORCE) learning and innate training}
We used two RC techniques called \textit{first-order-reduced and controlled-error (FORCE) learning} \cite{sussillo2009generating} and \textit{innate training} \cite{laje2013robust}.
Both FORCE learning and innate training are methods that harness the chaoticity of the system.
Below, we briefly describe the algorithms of both FORCE learning and innate training.

FORCE learning is a method that embeds designated dynamics in a system by harnessing the chaoticity of dynamical systems.
Suppose the following ESN dynamics with a single feedback loop:
\begin{align}
    \tau \diff{\bm{x}}{t}{}(t) &= -\bm{x}(t) + \tanh\left(gJ \bm{x}(t) + \bm{u}z(t)\right) \\
    z(t) &= \bm{w}^T\bm{x}(t)
\end{align}
Typically, the scaling parameter $g$ is set to be greater than $1$ to make the whole system chaotic \cite{sompolinsky1988chaos}.
In FORCE learning, to embed the target dynamics $f(t)$ in the system, $\bm{w}$ is trained to optimize the following cost function $C_{\text{FORCE}}$:
\begin{align}
    C_{\text{FORCE}} := \braket{\|z(t)-f(t)\|^2}
\end{align}
Here, the bracket denotes the averaged value over several samples and trials.
Especially in the FORCE learning, $\bm{w}$ is optimized online with a least-square error algorithm.
It was reported from numerical experiments using ESN that better training performance was obtained when the initial RNN was in a chaotic regime \cite{sussillo2009generating}.

Innate training is also a scheme for harnessing chaotic dynamics and is accomplished by modifying the internal connection $J$ using FORCE learning.
The novel aspect of innate training is that the inner connection of ESN is trained in a semi-supervised manner, that is, the connection matrix $J$ of the ESN is modified to minimize the following cost function $C_{\text{innate}}$ to reproduce the chaotic dynamics yielded by the initial chaotic RNN ($\bm{x_{\text{target}}}(t)$, \textit{innate trajectory}):
\begin{align}
    C_{\text{innate}}:=\braket{\|\bm{x}(t)-\bm{x}_{\text{target}}(t)\|^2}
\end{align}
Intriguingly, the innate trajectory is reproducibly generated for a certain period with the input while maintaining the chaoticity after the training.
In other words, innate training is a method that allows a chaotic system to reproducibly yield the innate trajectory with complicated spatiotemporal patterns by applying the FORCE learning method to the modification of the internal connection.

In this study, we propose a novel method of designing CI by employing both FORCE learning and innate training techniques.

\subsection{Recipe for designing chaotic itinerancy}
Our proposed method is a batch-learning scheme consisting of the following three-step process (\cref{fig:1}C).

\subsubsection{Designing quasi-attractor}
In step 1, the connection matrix $J^\text{ch}$ of the chaotic ESN is adjusted by innate training to design the trajectories of quasi-attractors.
First, the target trajectories $\bm{x}^{s}_{\text{target}}(t)$ are recorded for $M$ symbols under an initial connection matrix $J^{\text{init}}$ and some initial states $\bm{x}^{s}_{\text{target}}(0)$, where $\bm{x}^{s}_{\text{target}}(t)$ denotes chaotic dynamics when the symbol is switched to $s$ at $t=0$ ms (for simplification, the switching time is fixed to $t=0$ ms. in step 1. Note that the symbol can be switched at any time).
In step 1, $J^{\text{ch}}$ is trained to optimize the following cost function $C_{\text{1-in}}$:

\begin{align}
    C_{\text{1-in}} := \sum_{s\in S}\int_{0}^{L_{\text{innate}}} \|\bm{x}^{s}(t)-\bm{x}^{s}_{\text{target}}(t) \|^2 dt
\end{align}
Here, $\bm{x}^s(t)$ represents the dynamics when the symbol is switched to $s$ at $t=0$ ms, and $L_{\text{innate}}$ the time period of the target trajectory.
We only modify half the elements of $J^\text{ch}$.
The connection matrix $J^{ch}$ is trained for 200 epochs for each $s$.
We finally use $J^\text{ch}$ recording the minimum $C_{\text{1-in}} $ (see the Appendix for the detailed algorithm used in step 1).
After the innate training in step 1, the system is expected to reproduce the recorded innate trajectories $\bm{x}^{s}_{\text{target}}$ for $L_{\text{innate}}$.

Similarly, $\bm{w}_{\text{out}}$ is trained to produce designated output dynamics $f^{s}(t)$ corresponding to symbol $s$.
The following cost function $C_{\text{1-out}}$ is optimized:
\begin{align}
    C_{\text{1-out}} := \sum_{s\in S}\int_{0}^{L_{\text{out}}} \|f^{s}(t)-\bm{w}_{\text{out}}^T\bm{x}^{s}(t) \|^2 dt
\end{align}
Here, note that $L_{\text{innate}}$ does not always match $L_{\text{out}}$, that is, $L_{\text{out}}$ can be greater than, $L_{\text{innate}}$.
The training is accomplished by an offline algorithm Ridge regression based on the recorded internal dynamics $\bm{x}^{s}(t)$.

\subsubsection{Embedding autonomous symbol transition}
In step 2, we tune a feedback loop $f_{\text{softmax}}$ to achieve the autonomous symbol transition.
We especially prepare target periodic transition rules switching every $T$ [ms].
Suppose a target periodic symbol dynamics $s_{\text{per}}(t)$.
First, the network dynamics $\bm{x}(t)$ of the open-loop setup (\cref{fig:1}B(1)) is recorded with a symbol dynamics $s_{\text{per}}(t)$ for $T_{\text{rec}}:=500,000$ ms.
Based on the recorded dataset, $f_{\text{softmax}}$ is tuned to output $s_{\text{per}}(t)$ from $\bm{x}(t)$.
The parameters $\bm{w}_{s}$ of $f_{\text{softmax}}$ is trained to optimize the following cost function $C_{2}$:
\begin{align}
    C_{2} := -\sum_{s\in S}\int_{0}^{T_{\text{rec}}} \mathbbm{1}\left\{s_{\text{per}}(t) = s \right\} \log \frac{e^{\bm{w}_{s}^T \bm{x}(t)}}{\sum_{k \in S} e^{\bm{w}_{k}^T \bm{x}(t)}} dt
\end{align}
As the optimization algorithm, we use the limited-memory Broyden–Fletcher–Goldfarb–Shanno (BFGS) algorithm\cite{byrd1995limited}.

\subsubsection{Embedding stochastic symbol transition}
In step 3, we implement a stochastic transition rule governed by a finite state machine by modifying a feedback loop $f_{\text{softmax}}$.
As discussed in the Introduction, the chaoticity of the system is expected to be employed to emulate the stochastic process in the deterministic setup.
The process of the learning is same as that in step 2, that is, the pair of $(\bm{x}(t), s_{\text{sto}}(t))$ recorded in the open-loop setup for 500,000 ms is used to train the $f_{\text{softmax}}$ to emulate $s_{\text{sto}}(t)$.
Here, we use the following cost function $C_{3}$ in the training:
\begin{align}
    C_{3} := -\sum_{s\in S}\int_{0}^{T_{\text{rec}}} \mathbbm{1}\left\{s_{\text{sto}}(t) = s \right\} \log \frac{e^{\bm{w}_{s}^T \bm{x}(t)}}{\sum_{k \in S} e^{\bm{w}_{k}^T \bm{x}(t)}} dt
\end{align}
As with the optimization of the cost function $C_2$, $C_3$ is optimized with the Limited-memory BFGS algorithm.

\section{Results}
In this section, we show the demonstration and analytic results of the numerical experiments for each step.
\begin{figure*}[hbtp]
  \centering
  \includegraphics[width=0.85\textwidth]{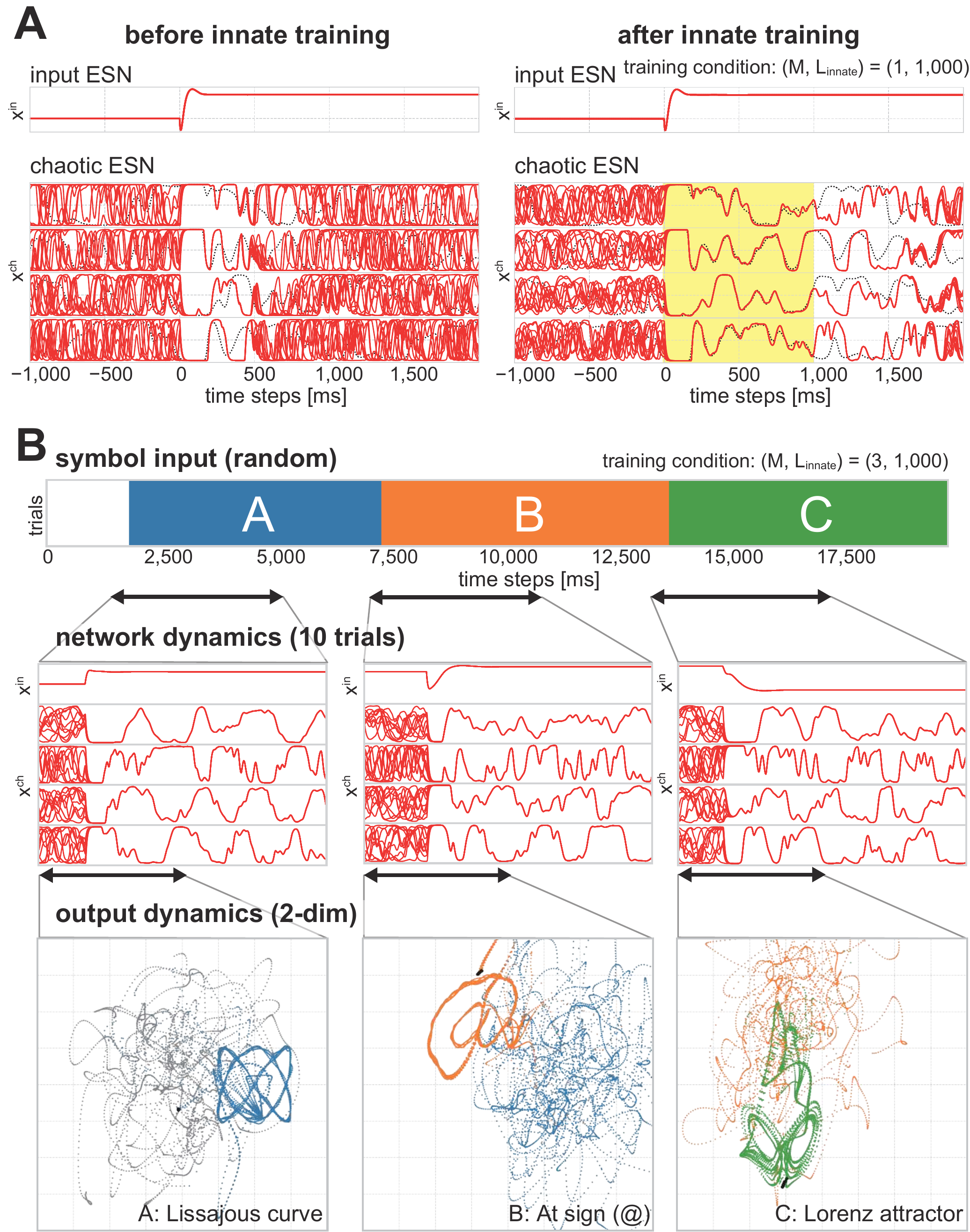}
  \caption[fig1]{
    Demonstration of step 1.
    (A) The dynamics of the reservoir before and after the innate training.
    In the figure, we show the RNN dynamics trained under the condition $(M, L_{\text{innate}})=(1, 1,000)$.
    The time-series data of a selected node in the input ESN is shown in the top column.
    Conversely, the four selected dynamics of the chaotic ESN are displayed in the bottom four columns.
    In each column, both the innate trajectory (black dotted) and ten individual trajectories with different initial conditions (red) are exhibited.
    (B) Demonstration of open-loop dynamics.
    The network dynamics of the RNN trained under the condition $(M, L_{\text{in}}, L_{\text{out}})=(3, 1,000, 1,500)$ is used in this demonstration.
    Both the network dynamics and output dynamics of the trained readout are depicted.
    The readout is trained to output the Lissajous curve for symbol A, at the sign for symbol B, and the xz coordinates of the Lorenz attractor for symbol C.
    Note that the intervals of the symbol input were randomly decided.
  }\label{fig:2}
\end{figure*}

\begin{figure*}[hbtp]
  \centering
  \includegraphics[width=0.85\textwidth]{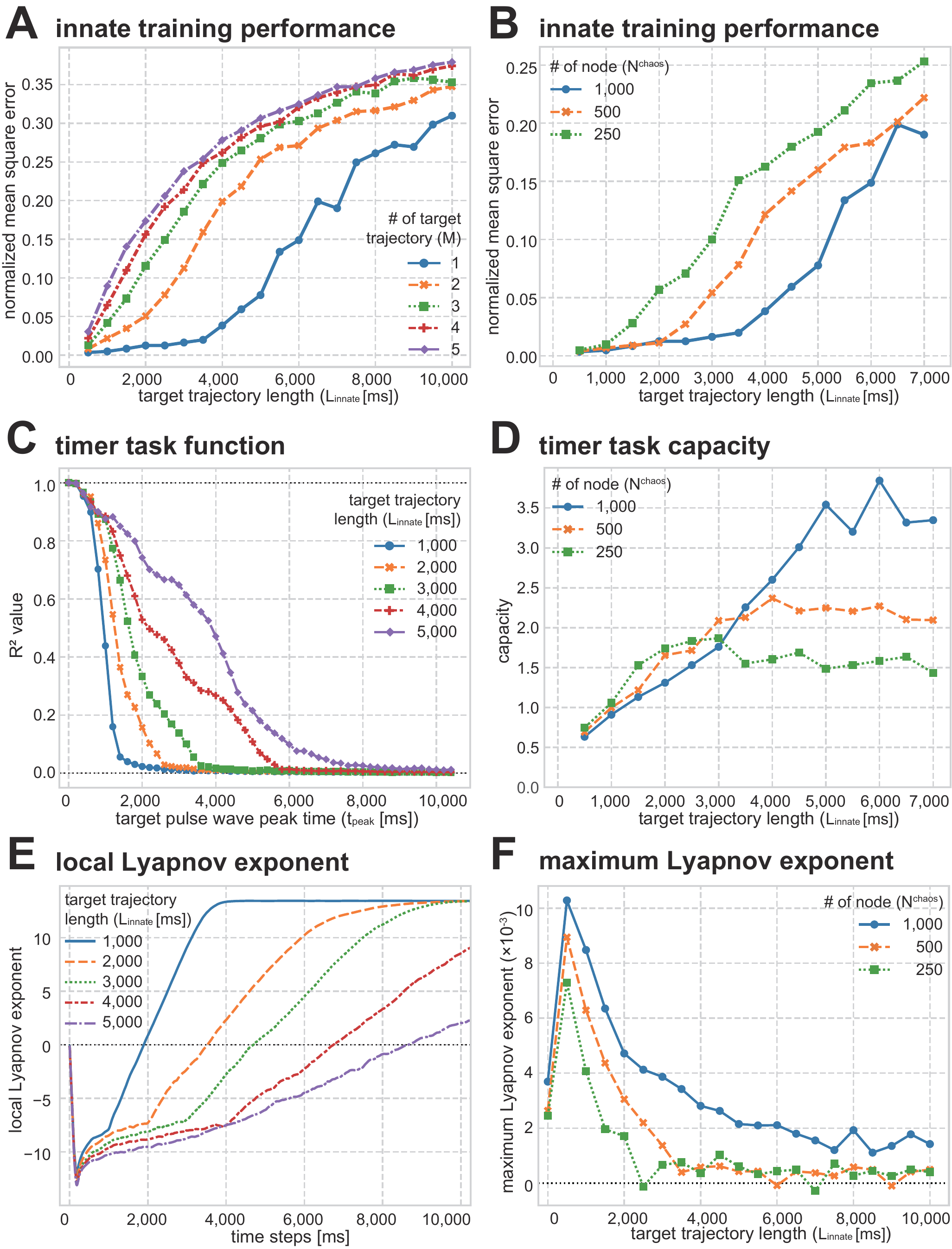}
  \caption[fig3]{
    (A) Performance of innate training over $M$ symbols.
    The normalized mean square errors (NMSEs) are calculated from the ten trials.
    (B) Effect of network size $N^{\text{ch}}$ on the performance of innate training.
    (C) Evaluation of the temporal information capacity with timer task.
    The averaged values for ten trials are plotted.
    (D) Effect of the system size on timer task capacities.
    Timer task capacity is defined as the integral value of the timer task function.
    (E) Evaluation of the local Lyapunov exponent (LLE).
    The LLE is measured with the time development of the perturbation of the chaotic ESN (see the Appendix for detailed information about the calculation method of the LLE).
    (F) Evaluation of the system's maximum Lyapunov exponent (MLE).
}\label{fig:3}
\end{figure*}

\renewcommand\thesubsection{Step \arabic{subsection}}
\subsection{Designing quasi-attractor}
As discussed in the previous section, the internal connection of the chaotic ESN $J^{\text{ch}}$ is trained to output the corresponding innate trajectories reproducibly to the symbol switching.
\cref{fig:2}A demonstrates the change of the network dynamics of a 1,500-node RNN ($N^{\text{in}}=500,~N^{\text{ch}}=1,000$) whose connection matrix is modified with innate training under the condition $(M, L_{\text{innate}})=(1,~1,000)$.
The trajectory quickly spreads before $t=L_{\text{innate}}$ in the pre-trained system, whereas the target trajectory $\bm{x}^{\text{target}}_s$ (dashed line) is reproducibly yielded for 1,000 ms (covered by the yellow rectangle) in the post-trained system.
Moreover, intriguingly, the dispersion of the trajectories continues to be suppressed even after $t=L_{\text{innate}}$.

Next, \cref{fig:2}B displays both the network dynamics and the output dynamics.
The 1,500-node RNN ($N^{\text{in}}=500,~N^{\text{ch}}=1,000$) trained under the condition $(M,L_{innate})=(3,~1,000)$ was used.
At first, the symbol input was absent, and then symbols were switched with random intervals from the middle.
Also, the 2-dim readout was trained to output the Lissajous curve for symbol A, at the sign for symbol B, and the xz coordinates of the Lorenz attractor for symbol C for $L_{\text{out}}=1,500$ ms.
It was observed that the desired spatiotemporal patterns were stably and reproducibly generated for a certain period in every trajectory with different initial states after the symbol transition (see supplementary video 1).
Note that the same linear model $\bm{w}_{\text{out}}$ was used in the demonstration, implying that the trajectory of each quasi-attractor has rich enough information to output the designated time-series patterns independently even with the single linear regressor.
Our scheme for designing transient dynamics would be highly useful in the field of robotics because the process in step 1 is easily achieved by adjusting the partial elements of a high-dimensional chaotic system.
For example, the system working in a real-world environment should immediately and adaptively switch its motion according to the change of environmental input like a system developed by Ijspeert et al. \cite{ijspeert2007swimming}, which can be easily accomplished by our computationally cheap method.
In this way, our method would work effectively in the context of robotics, where fast responsiveness and adaptability are required.

We also examined both the scalability and the validity of innate training in detail through several numerical experiments (\cref{fig:3}).
First, we examined the relationship between the number of input symbols $M$ and the accuracy of innate training.
To evaluate the performance of innate training, we used the normalized mean square error (NMSE) between the output and the innate trajectory $\bm{x}_{\text{target}}^{s}$ represented by the following formula:
\begin{align}
    NMSE := \frac{1}{M} \sum_{s\in S} \left< \frac{\int_{0}^{L_{\text{innate}}} \|\bm{x}^{s}(t)-\bm{x}^{s}_{\text{target}}(t) \|^2 dt}{\int_{0}^{L_{\text{innate}}} \|\bm{x}^{s}_{\text{target}}(t) \|^2 dt} \right>
\end{align}
We calculated the NMSE for ten trials.
\cref{fig:3}A shows the innate training performances with the different training conditions, suggesting that NMSEs are more likely to increase with a longer target trajectory and a larger number of symbols.
This result implies that innate training has its limitation in the design of the quasi-attractors.
We also examined the effect of network size on the capability to embed the quasi-attractors.
We investigated the relationship between the number of nodes in the chaotic ESN $N_c$ and the accuracy of innate training under the condition $M=1$ ( \cref{fig:2}B), suggesting that the NMSEs were less likely to increase with a larger network.
To summarize, our analysis indicates that longer trajectories can be embedded in a larger network by innate training.

\begin{figure*}[hbtp]
  \centering
  \includegraphics[width=0.8\textwidth]{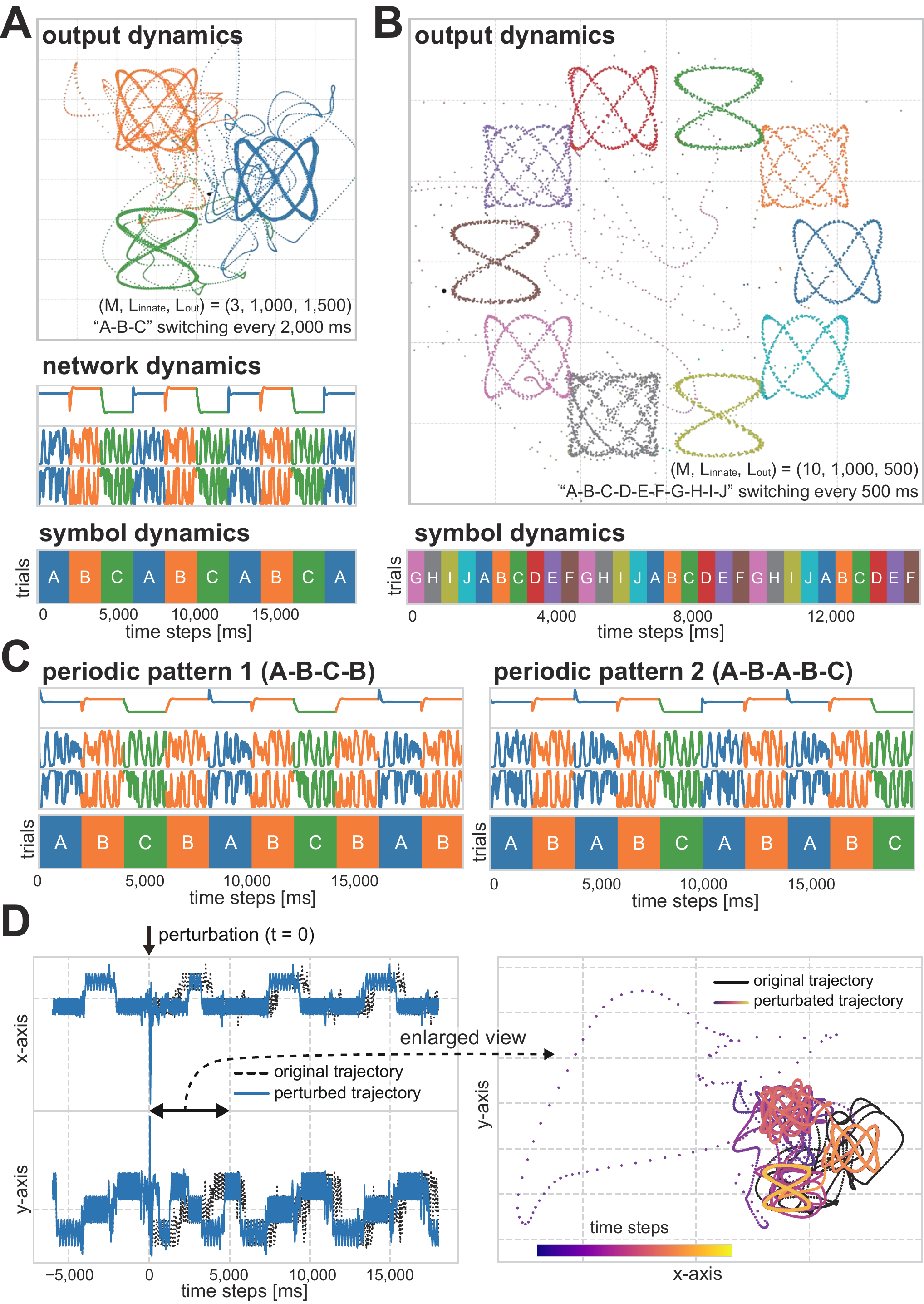}
  \caption[fig4]{
    Demonstrations of closed-loop dynamics in step 2.
    (A) Three-symbol periodic transition.
    We prepared an RNN trained under the condition $(M,~L_{\text{innate}})=(3,~1,000)$ and a readout trained under the condition $L_{\text{out}}=1,500$ ms to output three Lissajous curves corresponding to the symbol input.
    The feedback loop $f_{\text{softmax}}$ realizes the periodic symbol transition A-B-C switching at 2,000-ms intervals.
    (B) Ten-symbol periodic transition.
    We prepared an RNN trained under the condition $(M,~L_{\text{innate}})=(10,~500)$ and a readout trained under the condition $L_{\text{out}}=500$ ms to output ten different Lissajous curves corresponding to the symbol input.
    The feedback loop $f_{\text{softmax}}$ achieves the periodic symbol transition A-B-C-D-E-F-G-H-I-J switching at 500-ms intervals.
    (C) Demonstration of the tasks requiring higher-order memory to be solved.
    The same RNN was used in the demonstration of (A).
    The left panel displays the periodic symbol transition pattern A-B-C-B switching at 2,000-ms intervals.
    The right one demonstrates the periodic symbol transition pattern A-B-A-B-C switching at 2,000-ms intervals.
    These tasks were accomplished in the same way in demonstrations (A) and (B), that is, only the parameters in $f_{\text{softmax}}$ were tuned.
    (D) Two output dynamics: original trajectory and perturbed trajectory.
    A small perturbation was given to the original trajectory at $t=0$ ms.
  }\label{fig:4}
\end{figure*}

Next, we evaluated the effect of innate training on the capacity of the system's information processing.
We prepared a \textit{timer task} and measured how long the inputted information was stored in the RNN.
In the timer task, the pulse-like wave with a peak $t_{\text{peak}}$ [ms] after the symbol transition was prepared as the target, and the performance was defined as the accuracy of the pulse-like wave reconstruction by a trained readout. Here, we defined the $R^2$ value between the output and the pulse-like wave as the \textit{timer task function} $R^2(t_{\text{peak}})$.
At the same time, we also calculated the integral value of the timer task function $\int_{0}^{\infty} R^2(t) dt$ and define it as the \textit{timer task capacity} (see the Appendix for detailed information about the setup of the timer task).
\cref{fig:4}C shows the timer task function with different innate training conditions, indicating that RNNs trained with the longer-length target trajectory $L_{\text{innate}}$ perform better.
It was also observed that the timer task capacity saturated around $L_{\text{innate}}=5,000$ ms in the 1,000-node RNN, and the border of the saturation decreased in a smaller system (\cref{fig:3}D).
These results imply that the temporal information capacity of the system is improved by innate training with the longer target length $L_{\text{innate}}$ but saturates at a certain value, which is determined by the system size.

Furthermore, we assessed the effect of innate training on the system's chaoticity by measuring the Lyapunov exponents of the system.
Since the transition among quasi-attractors is driven by the system's chaoticity, it is necessary to keep the system chaotic.
In this experiment, we measured the local Lyapunov exponent (LLE) to evaluate the degree of trajectory variation after the symbol switching.
We also measured the maximum Lyapunov exponent (MLE) without any inputs ($\bm{u}(t)=\bm{0}$) to estimate the global chaoticity of the system (see the Appendix for the detailed calculation algorithm of both the LLE and MLE).
\cref{fig:4}E displays the LLE values of the systems with the different target trajectory length $L_{\text{innate}}$, suggesting that the trajectories unevenly expand after the symbol transition.
In particular, it was observed from the LLE analysis that contracting regions existed (regions with negative LLEs corresponding to the lengths of the quasi-attractors) caused by the transient dynamics projected by the input ESN, and the degree of the expansion became gradual in the trained period $t\in[0, L_\text{innate})$.
These results imply that innate training yields a locally contractive phase space structure, that is, a quasi-attractor. Moreover, positive MLE values were constantly obtained from the MLE analysis depicted in \cref{fig:3}F, supporting the conjecture that the system chaoticity was maintained especially well with the larger RNNs even after the innate training. (Note that a sharp increase in MLE was observed with shorter $L_{\text{innate}}$, which is caused by the increase in the spectral radius of the connection matrix $J$ of the system. See the Appendix for detailed information of the analysis.)

\subsection{Periodic symbol transition}
In step 2, the system autonomously generates a symbol sequence externally given in step 1.
The additional feedback loop realizes the autonomous periodic switching of the symbols.
We demonstrate that various types of periodic symbol sequences switching at a fixed interval can be easily designed simply by modifying the parameter of the feedback loop $f_{\text{softmax}}$.
\cref{fig:4}A demonstrates the embedding of the periodic symbol sequence A-B-C (2,000-ms interval and 6,000-ms period) with a trained RNN ($(M,L_{\text{innate}})=(3,~1,000)$).
\cref{fig:4}A also exhibits the embedding of the periodic symbol sequence A-B-C-D-E-F-G-H-I-J (500-ms interval and 5,000-ms period), with the same RNN used as the demonstration in \cref{fig:4}A.
In both demonstrations, the system succeeded not only in generating the desired symbol transition rules but also in stably outputting the designated output dynamics with high accuracy.

We also show that the system can solve tasks requiring higher-order memory in the same scheme.
We prepared the two periodic symbol sequences A-B-C-B and A-B-C-B-A.
These two symbol sequences are more difficult to embed because the system must change the output according to the previous output.
In the symbol transition A-B-C-B, for example, the system must output the next symbol depending on the previous symbol when switching from B, though the total number of symbols is the same as in the task A-B-C.
We used the same RNN and setup used in the \cref{fig:4}A and only changed the parameters in $f_{\text{feedback}}$ to realize the symbol transitions.
\cref{fig:4}C displays the network dynamics and symbol transition of the two tasks, showing that the system successfully achieves both the periodic sequence A-B-C-B with an 8,000-ms period and A-B-A-B-C with a 10,000-ms period.
These results suggest that the trained RNN had the higher-order memory capacity, that is, the generated trajectories have sufficient separability to distinguish the contextual situation depending on the previous symbol sequence (see supplementary video 2).
In robotics, periodic motion control has often been implemented by an additional oscillator (e.g., a central pattern generator) to yield limit cycles \cite{ijspeert2007swimming,steingrube2010self,liu2013central,owaki2013simple}.
Our method in step 2 would be useful in designing limit cycles with longer periods and more complicated patterns.

We also analyzed the effect of perturbation to investigate the stability of the embedded symbol transition.
\cref{fig:4}D shows the output dynamics of both the original and perturbed trajectories, clarifying that the trajectory returned to the original one after the addition of the perturbation.
We also calculated the MLE values of the system and obtained the value $-1.89\times10^{-4}$, which was very close to zero.
These analyses indicate that the trained feedback loop $f_{\text {innate}}$ made the system non-chaotic, that is, the generated internal dynamics was a limit-cycle.

\begin{figure*}[hbtp]
  \centering
  \includegraphics[width=0.8\textwidth]{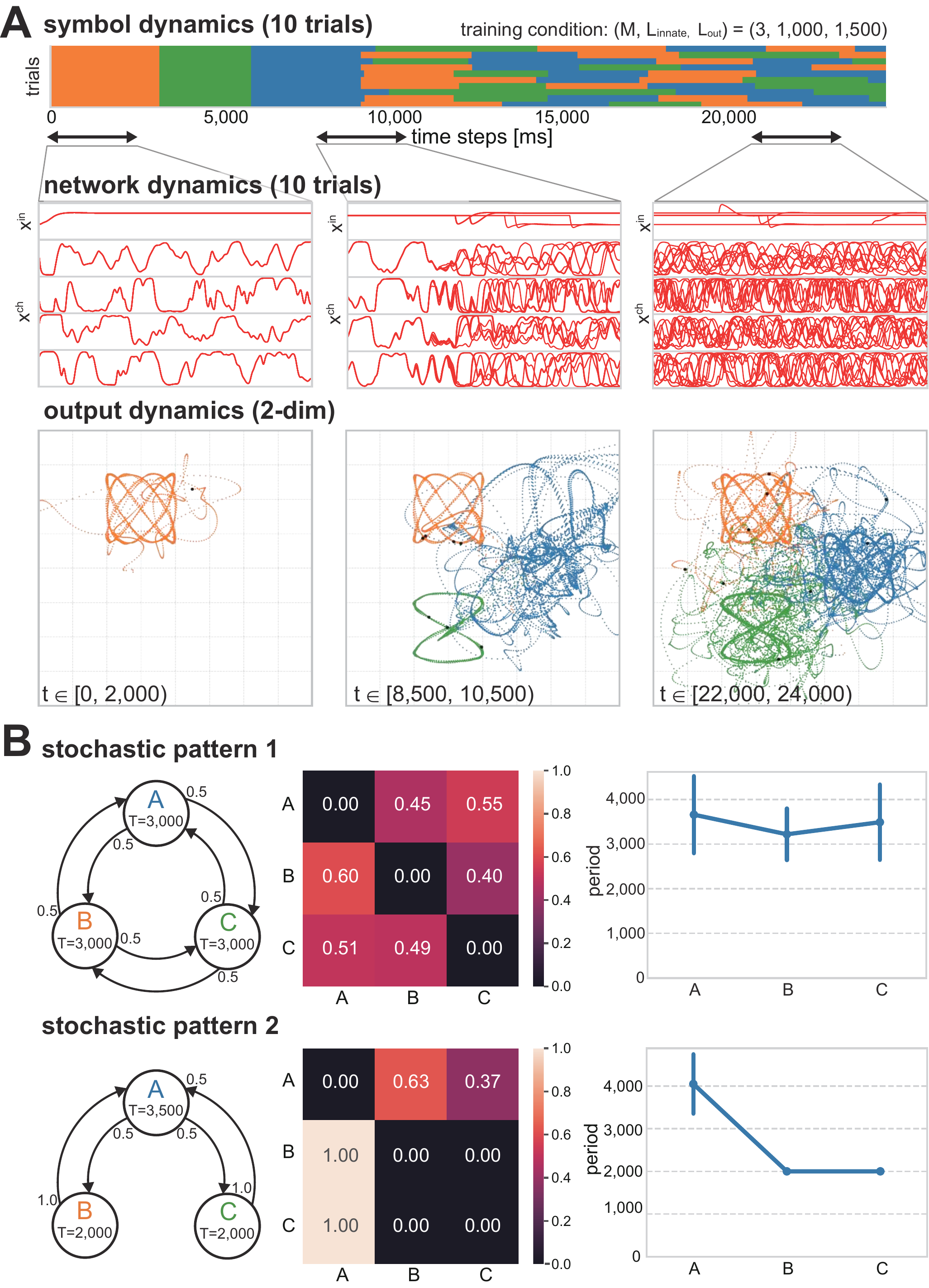}
  \caption[fig5]{
    Demonstration of step 3.
    (A) Network dynamics with $f_{\text{feedback}}$ trained to imitate a stochastic transition rule.
    We used an RNN trained under the condition $(M, L_{\text{innate}})=(3, 1,000)$, and readout trained to output Lissajous curves under the condition $L_{\text{out}}=1,500$ ms.
    The feedback classifier $f_{\text{softmax}}$ was trained to uniformly switch the symbol among the three symbols A, B, and C at 3,000-ms intervals.
    Ten different trajectories with small perturbations are overwritten in the figure.
    (B) Evaluation of the embedding performance of a stochastic symbol transition.
    Two different stochastic symbol transition rules (patterns 1 and 2) were prepared as the target.
    The same RNN was used as in the demonstration of (A).
    The middle figures show the obtained probability density matrix, and the right ones show the average switching duration (the error bar represents standard deviation).
  }\label{fig:5}
\end{figure*}

\begin{figure*}[hbtp]
  \centering
  \includegraphics[width=0.85\textwidth]{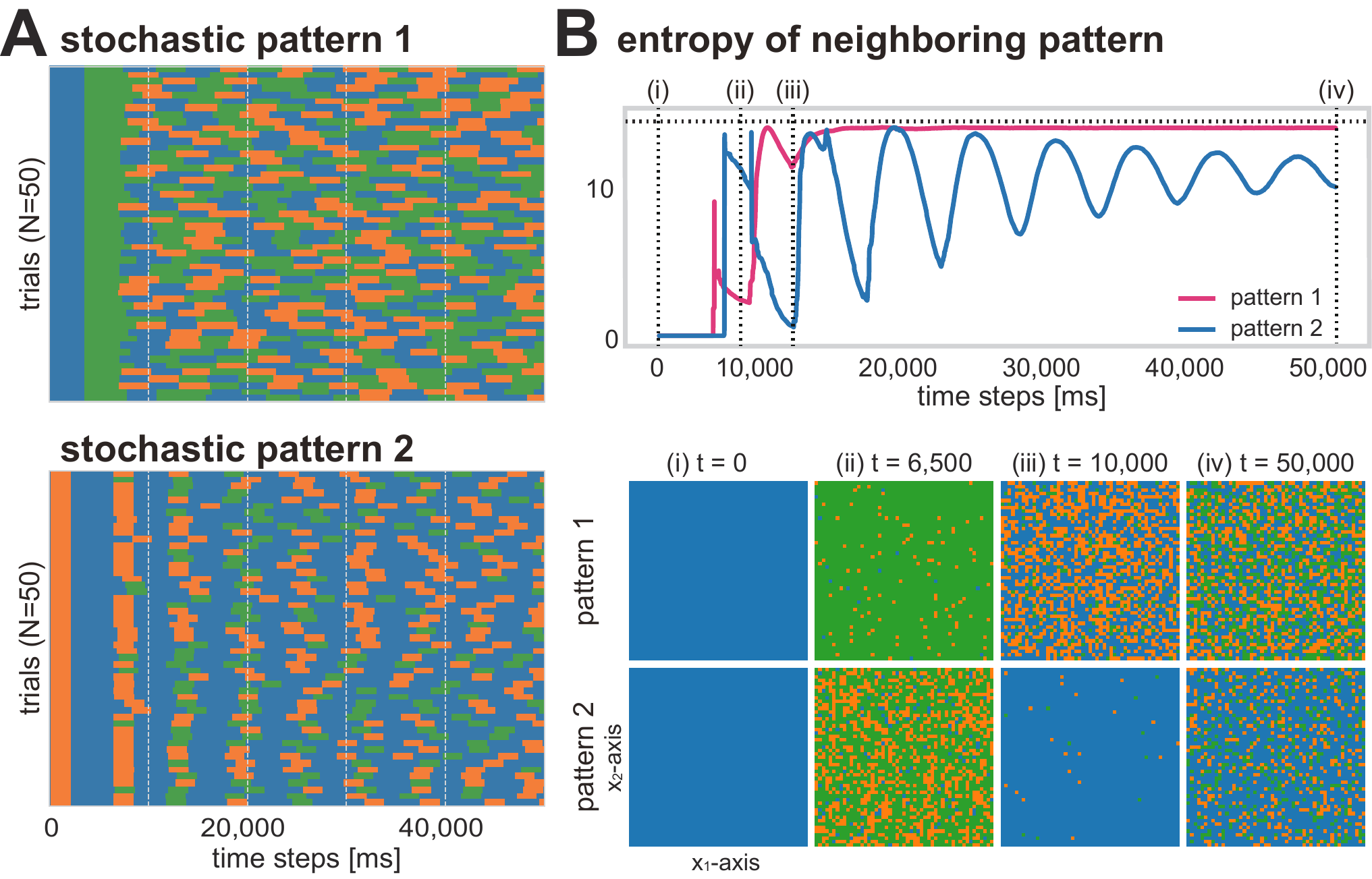}
  \caption[fig6]{
    Analysis of symbol dynamics and the final state.
    (A) Effect of a small perturbation on the terminal symbol dynamics.
    We evaluated the two closed-loop set-ups prepared in \cref{fig:5}B.
    The figures display the symbol dynamics generated by 50 trajectories with 50 different initial values.
    (B) Analysis of symbol dynamics generated by the temporal development of the initial states on a small plane and its entropy of the symbol pattern.
    Two dimensions ($x_1, x_2$) on the phase space were selected from the chaotic ESN to construct the plane.
    We observed the symbol dynamics generated by the temporal development of the states on the plane.
    To evaluate the randomness of the obtained pattern, we calculated the entropy of the obtained symbolic pattern based on the probability distribution constructed from the $3\times3$ grid patterns.
    Note that the horizontal dotted line shows the maximum entropy ($\log_2{3^9}\approx 14.26$).
  }\label{fig:6}
\end{figure*}

\subsection{Stochastic symbol transition (chaotic itinerancy)}
In step 1, we constructed the trajectories of the quasi-attractors and the corresponding output dynamics.
In step 2, we showed that periodic transitions among quasi-attractors can be freely designed by simply tuning the feedback loop $f_{\text{feedback}}$.
In step 3, we realize a stochastic transition, that is, CI.
As discussed above, the system is expected to employ its chaoticity to emulate a stochastic transition in deterministic dynamical systems.

First, we demonstrate that stochastic transition can be freely designed by adjusting $f_{\text{feedback}}$ (see \cref {fig:5}A and supplementary video 3).
In this demonstration, we used the same RNN as in \cref{fig:4}A.
We prepared a symbol transition rule uniformly switching among symbols A, B, and C at 3,000-ms intervals.
\cref{fig:5}B shows the symbol dynamics, network dynamics, and output dynamics, suggesting that the symbol transitions started to spread at around $t=10,000$ ms and finally settle down to completely different transition patterns.
Nevertheless, the system continued to generate Lissajous curves stably.
These demonstrations imply that the system constantly reproduced quasi-attractors embedded by innate training, while the quasi-stochastic transition was achieved by the global chaoticity.

To analyze the flexibility of our method, we measured the stochastic transition matrix and the average symbol intervals (\cref{fig:5}B).
We prepared two stochastic symbol transition rules as the targets: the transition rule governed by the uniform finite state machine (\textit{pattern 1}) and the transition rule governed by the finite state machine with a limited transition (\textit{pattern 2}).
Note that we used the same trained RNN as in the demonstration in step 2 and embedded the transition rules simply by adjusting $f_{\text {softmax}}$.
\cref{fig:5}B shows the results of the obtained trajectories, implying that the system successfully embedded patterns similar to the target rules, although there were some errors and variations in the transition probability and the switching time.
The positive MLEs were obtained in both cases ($+2.01\times 10^{-3}$ in pattern 1, and $+1.71\times 10^{-3}$ in pattern 2), suggesting that the system was weakly chaotic as a whole.

Finally, we analyzed both the structures of the obtained chaotic attractors and the symbol dynamics in detail.
\cref{fig:6}A shows the effect of small perturbations on the symbol dynamics, implying that the patterns of symbol dynamics varied after a certain period.
To analyze the structural change of the terminal symbol state, we measured the symbol dynamics accompanied by the temporal development of the set of initial states on a plane constructed by the two selected dimensions (\cref{fig:6}B), clarifying that a complex terminal symbol structure emerges after a certain period (\cref{fig:6}B).
Especially in the embedding of the pattern 1 rule, the entropy of the terminal symbol pattern converges to a value close to the maximum entropy $\log_2{3^9}\approx 14.26$ (note that the entropy was measured based on the probability distribution constructed by the frequency of $3\times 3$ grid patterns).
These results indicate that the symbol transition dramatically changed even with a small perturbation and was unpredictable after a certain period, that is, the prediction of symbol dynamics required the complete observation of the initial state value and calculation of the temporal development with infinite precision.

\section{Discussion}
In this study, we proposed a novel method of designing CI based on RC techniques.
We also showed that the various types of output dynamics and symbol transition rules could be designed with high operability simply by adjusting the partial parameters of a generic chaotic system with our three-step recipe.
In this section, we discuss the scalability of our method and the mechanism of how CIs are successfully embedded by reviewing several numerical analyses that verify the validity of innate training.

First, the results of the innate training performances displayed \cref{fig:3}A and B indicate that the number of RNN nodes constrains the total length of the quasi-attractors that can be embedded in the system by the innate training.
However, the LLE analyses in \cref{fig:3}E show that the system has the expanded region of the negative LLE even when the NMSE between the innate trajectory and the embedded trajectory becomes large (e.g., $ L_{\text{innate}}=5,000$ ms).
These results imply that even when the innate trajectories are not successfully embedded in the system, the system stably yields high-dimensional trajectories with complicated spatiotemporal patterns for each symbol transition over $L_{\text{innate}}$, caused by the weakening of the system chaoticity.
In fact, the same RNN trained under the condition $(M,L_{\text{innate}})=(3,~1,000)$ was repeatedly used in our series of demonstrations, the desired output dynamics (e.g., the Lissajous curves) being constantly generated for $L_{\text{out}}=1,500$ ms periods after the symbol shift (\cref{fig:2,fig:4,fig:5}).
We also demonstrated that the system can autonomously generate a symbol transition rule with an interval greater than $L_{\text{innate}}$ (\cref{fig:4,fig:5}), suggesting that the system exploited high-dimensional reproducible trajectories longer than $L_{\text{innate}}$.

Moreover, it is assumed that the length of the quasi-attractors constrains the target stochastic transition rules that can be embedded.
Indeed, the system failed to imitate the stochastic transition, and the transition became periodic when the target transition had a shorter switching interval, whereas the training of $f_{\text{feedback}}$ became unstable when it had a longer switching interval.
These results suggest that the following two mechanisms should be required in the design of CI in our method:
(i) the differences among the trajectories are sufficiently enlarged through the temporal development to realize the stochastic symbol transition, and
(ii) a similar spatiotemporal pattern should be reproducibly yielded until the switching moment to discriminate the switching timing precisely.
These two mechanisms are contradictory, of course, and the desired CI can likely be embedded when both conditions are moderately satisfied.

Next, we discuss the effectiveness and significance of our method from the viewpoint of robotics.
It should be noted that both the trajectory of the quasi-attractor and its transition rule are freely designed in a high-dimensional chaotic system by adjusting the reduced number of parameters and utilizing the intrinsic high-dimensional chaos.
This flexibility is unavailable in the conventional heuristic methods of CI, and our method offers a novel methodology for designing CI on a wider range of chaotic dynamical systems, including real-world robots.
Our cheap training method would also be useful in designing more sophisticated agents, such as animals.
For example, recent physiological studies on the motor cortex \cite{stroud2018motor, perich2018neural} support the conjecture that a large variety of behaviors can be instantaneously generated by the partial plasticity of the nervous system, indicating that adjustments of entire neural circuits are not necessary.
In this point, our computationally inexpensive method of reusing and fine-tuning the trained model would be especially helpful in the context of bio-inspired robotics, where fast responsiveness and the real-time processing are required.

Another advantage of our method is that it does not require the explicit structure of dynamical systems.
For example, in the method proposed by Namikawa and Tani \cite{namikawa2008model,namikawa2010learning,namikawa2011neurodynamic}, the controller needs a fixed hierarchical structure and modularity.
Therefore, the trained controller was specialized in implementing a specific behavior, making it difficult to divert it for any other purpose.
In contrast, we proposed a method of designing CI with a generic setup consisting of a single chaotic ESN, auxiliary symbols, and an interface between them (input ESN) with high scalability.
Thus, our method allows us to design the various trajectories and their transition rules in a single high-dimensional chaotic system and thus greatly expands the scope of application of high-dimensional chaotic dynamical systems.
Especially, neuromorphic devices based on physical reservoir computing framework would be an excellent candidate for implementing our scheme.
Sprintronics devices, for example, are recently shown to exhibit chaotic dynamics \cite{taniguchi2019chaos,yamaguchi2019synchronization} and are actively exploited as physical reservoirs \cite{torrejon2017neuromorphic,furuta2018macromagnetic,tsunegi2019physical}.
We expect that this framework would provide one of the promising application scenarios for real-world implementations of our scheme.

Our method is also scalable to autonomous symbol generation required in more advanced functionality.
For example, in our method, $M$ kinds of auxiliary symbols are given in advance.
However, in a highly autonomous system, such as humans, symbols are dynamically generated and destroyed due to developmental processes.
As demonstrated by Kuniyoshi et al. \cite{kuniyoshi2006early}, such self-organizing symbol dynamics can be realized by providing an additional automatic labeling mechanism in the system.
In other words, it is possible to generate symbols spontaneously by embedding an unsupervised learning algorithm in the system; this is a subject for future work.

Finally, the dynamic phenomena obtained by our method are significant from the viewpoint of high-dimensional dynamical systems.
As shown in \cref{fig:6}, we demonstrated that small differences in the initial network state were expanded by the chaoticity of the system, which eventually led to drastic change in both the global symbol transition pattern $s(t)$ and the local dynamics $\bm{x}(t)$.
Such tight interaction between micro-layer and macro-layer is a phenomenon unique to deterministic dynamical systems; that is, it cannot occur in principle in a system where the higher-order mechanism is completely separated from the lower-order one (e.g., independent random variables).
Also, the global characteristics of dynamical systems are often analyzed by the mean-field theory.
However, the analysis by the mean-field approximation cannot capture the contribution of microscopic dynamics to the macroscopic change.
Thus, our CI design method has a meaningful role in shedding light on the interaction between micro and macro dynamics in deterministic chaotic dynamical systems.

\begin{acknowledgments}
This work was based on results obtained from a project commissioned by the New Energy and Industrial Technology Development Organization (NEDO).
K.N was supported by JSPS KAKENHI Grant Numbers JP18H05472 and by MEXT Quantum Leap Flagship Program (MEXT Q-LEAP) Grant Number JPMXS0118067394.
\end{acknowledgments}
\bibliography{main}

\appendix
\setcounter{figure}{0}
\renewcommand{\thefigure}{S\arabic{figure}}

\begin{algorithm}[H]
  \caption{Innate training}
  \begin{algorithmic}[1]
    \For{$i \in A$}
        \State $P^i \leftarrow I$
    \EndFor
    \For{$s \in S$} \Comment{generating innate trajectory}
        \State initializing $\bm{x}^{s}_{\text{target}}(0)$  \Comment{washing out}
        \State recording $\bm{x}^{s}_{\text{target}}(t)~(t\in[0, L_{\text{innate}}))$
    \EndFor

    \For{$\text{epoch}=1$ to $200$} \Comment{innate training}
        \For{$s \in S$}
            \State initializing $\bm{x}^{s}(0)$ \Comment{washing out}
        \EndFor
        \State $t \leftarrow 0$
        \State $\text{count} \leftarrow 0$
        \While{$t<L_{\text{innate}}$}
            \For{$s \in S$}
                \If{$\text{count}=0~(\text{mod}~2)$}
                    \State continue \Comment{avoiding redundant sampling}
                \EndIf
                \State $\bm{e} \leftarrow \bm{x}^s(t) - \bm{x}^{s}_{\text{target}}(t)$
                \For{$i \in A$}
                    \For{$j \in B(i)$}
                        \State $J_{ij} \leftarrow J_{ij} - \bm{e}_i \sum_{k \in B(i)}P^i_{jk} \bm{x}^{s}_{k}(t)$
                        \For{$k \in B(i)$}
                            \State $Q^i_{jk} \leftarrow P^i_{jk} - \frac{\sum_{m \in B(i)}\sum_{n \in B(i)}P^i_{jm}\bm{x}^{s}_{m}(t)\bm{x}^{s}_{n}(t) {P^{i}_{nk}}}{1 + \sum_{m \in B(i)}\sum_{n \in B(i)}\bm{x}^{s}_{m}(t)P^{i}_{mn}(t)\bm{x}^{s}_{n}(t)}$
                        \EndFor
                    \EndFor
                    \State $P^i \leftarrow Q^i$
                \EndFor
            \EndFor
            \State $t \leftarrow t + \Delta t$
            \State $\text{count} \leftarrow \text{count} + 1$
        \EndWhile
    \EndFor
  \end{algorithmic} \label{algorithm:innate_training}
\end{algorithm}

\section{Algorithm of the innate training}
The connection matrix $J$ in step 1 is trained by algorithm shown in \cref{algorithm:innate_training}, where $A$ is a set of the selected nodes in the chaotic ESN ($|A|=N_c/2$), and $B(i)$ is a set of nodes projected to node $i$ among the nodes of the chaotic ESN.
In this study, $\Delta t = 1$ ms was constantly used.

\begin{figure*}[htbp]
  \centering
  \includegraphics[width=0.85\textwidth]{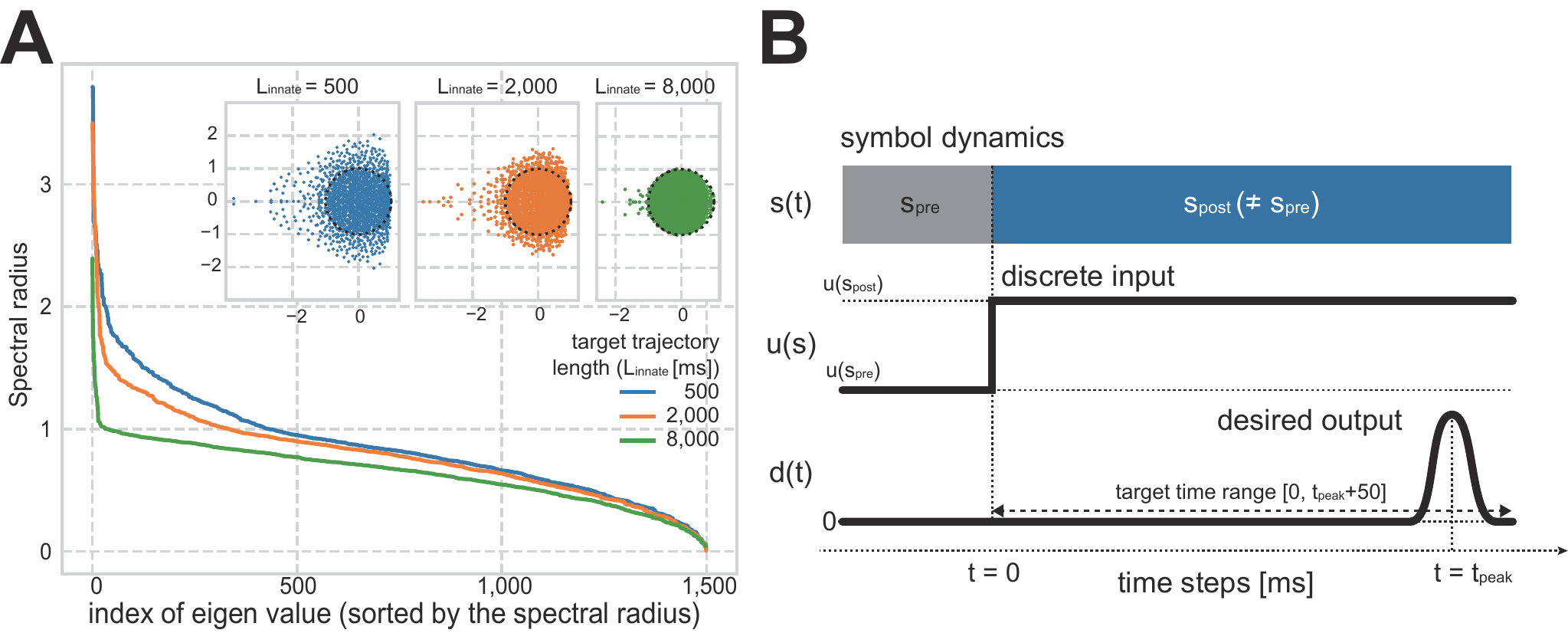}
  \caption[figs1]{
  (A) Distribution of eigenvalue and spectral radius of $J$ after the innate training.
  The results with three different length of the innate trajectories ($L_{\text{innate}}=500$ ms, $2,000$ ms, $8,000$ ms) are displayed.
  (B) Schematic graph of the timer task.
  The readout is trained to output a pulse wave with a delay $t_\text{peak}$ [ms] after symbol shift.
  }\label{fig:s1}

\end{figure*}

\section{Projected dynamics onto chaotic ESN}
To prevent chaotic ESN from becoming non-chaotic due to the bifurcation caused by the strong bias term, we tuned the feed-forward connection matrix $J^{\text{ic}}$ between input ESN and chaotic ESN before hand to project the following transient dynamics converging to $0$ onto the chaotic ESN when the same symbol input continues to be given:
\begin{align}
 J^{\text{ic}}\bm{x}^{\text{in}}(t-t_{\text{switch}}) \approx \left[(t-t_{\text{switch}}) \cdot \exp \left(-\frac{1}{2}\left(\frac{t-t_{\text{switch}}}{50}\right)^2\right)\right]\bm{v}_{s} \label{equ:connection}
\end{align}
where the elements of $\bm{v}_s \in \Real^{N^{\text{ch}}}$ (corresponding to symbol $s$) are also sampled from a normal distribution $\mathcal{N}(0,1)$.
This transition dynamics can be replaced with any other dynamics converging to $0$.

\section{Distribution of eigenvalues after the innate training}
\cref{fig:s1}A shows the effect of innate training on the distribution of eigenvalues under the condition $M=1$.
The spectral radius greatly exceeds $1$ when $L_{\text{in}}$ is small.
Conversely, the eigenvalues fit within the unit circle as $L_{\text {in}}$ becomes longer.
This result corresponds to the result in \cref{fig:3}F.

\section{Timer task setup}
The timer task evaluates the temporal information-processing capability of the system.
We evaluate how successfully the system can reconstruct a pulse-like wave with a peak with a delay $t_{\text{peak}}$ [ms] using only the linear transformation of the internal state $\bm{x}$ (\cref{fig:s1}B); that is, the readout $\bm{w}_{\text{out}}$ is trained to approximate a pulse-like wave represented by the following equation:
\begin{align}
     \bm{w}_{\text{out}}^T\bm{x}(t)  \approx \exp\left(\frac{(t-t_\text{peak})^2}{2\times 10^2}\right)
\end{align}
Note that $\bm{w}_{\text{out}}$ is tuned by Ridge regression with the Ridge parameter $\alpha=1.0$.
The task performance is calculated by averaging ten $R^2$ values between output and target, the $R^2$ value between $x$ and $y$ defined by $\frac{Cor(x,y)^2}{\sigma_{x}^2\sigma_{y}^2}$.
We express the task performance as $R^2(t_{\text {peak}})$ (timer task function).
Also, the timer task capacity is defined as the integral of $R^2(t_{\text{peak}})$ (i.e., $\int_{0}^{\infty} R^2(t)dt$).

\begin{algorithm}[H]
  \caption{Maximum Lyapunov exponent}
  \begin{algorithmic}[1]
    \State initializing $\bm{x}(0) \in \Real^{N}$ \Comment{reference trajectory}
    \State sampling $\bm{\epsilon} \in \Real^{N}$ ($\bm{\epsilon} \sim \mathcal{N}(0, 1)$)
    \State $\bm{\epsilon} \leftarrow l_{\text{pert}} \frac{\bm{\epsilon}}{\|\bm{\epsilon}\|}$ \Comment{normalizing sampled perturbation}
    \State initializing $\bm{y}(0) := \bm{x}(0) + \bm{\epsilon}$ \Comment{perturbed trajectory}
    \State $t \leftarrow 0$
    \State $\text{Lyap} = \{\phi\}$ \Comment{set of sampled MLE}
    \While{$t < T_\text{horizon}$}
        \State $\text{Lyap} \leftarrow \text{Lyap} \cup \left\{ \frac{1}{\Delta T} \log \left(\frac{\|\bm{y}(t+\Delta T)-\bm{x}(t+\Delta T)\|}{\|\bm{y}(t)-\bm{x}(t)\|}\right) \right\}$
        \State $t \leftarrow t + \Delta T$
        \State $\bm{y}(t) \leftarrow \bm{x}(t) + l_{\text{pert}} \frac{\bm{y}(t)-\bm{x}(t)}{\|\bm{y}(t)-\bm{x}(t)\|}$ \Comment{recomputing $y(t)$}
    \EndWhile
    \State calculating MLE by averaging values in $\text{Lyap}$
  \end{algorithmic} \label{algorithm:mle}
\end{algorithm}

\section{Maximum Lyapunov exponent}
MLE is calculated based on \cite {shimada1979numerical}.
The algorithm is represented in \cref{algorithm:mle}.
In this experiment, we used $\Delta T=1,000$ ms, $T_{\text{horizon}}=1,000,000$ ms,  $l_{\text{pert}}=10^{-6}$ and measured the average of ten trials.

\section{Local Lyapunov exponents}
LLE with symbol input $s$ is defined by the following formula:
\begin{align}
  \text{LLE}(t) := \left< \log \left( \frac{ \|\bm{x}^{s}_{\text{pert}}(t)-\bm{x}^{s}(t)\|}{\|\bm{x}^{s}_{\text{pert}}(0)-\bm{x}^s(0)\|} \right) \right>
\end{align}
where $\bm{x}^{s}$ represents the internal dynamics with the symbol input $s$ starting at time $t=0$ ms, $\bm{x}^{s}_{\text{pert}}$ represents the perturbed trajectories, a small perturbation being added to the original trajectory $\bm{x}^{s}$ at $t=0$ ms ($ \bm{x}^{s}_{\text{pert}}(0) = \bm{x}^{s}(0) + \bm{\epsilon}, \|\bm{\epsilon}\|=10^{-6}$).
The bracket represents an average over several trials.
We measured it with ten trials.
\end{document}